\documentclass{article}

\usepackage[preprint]{neurips_2023}

\usepackage[utf8]{inputenc} % allow utf-8 input
\usepackage[T1]{fontenc}    % use 8-bit T1 fonts
\usepackage{hyperref}       % hyperlinks
\usepackage{url}            % simple URL typesetting
\usepackage{booktabs}       % professional-quality tables
\usepackage{amsfonts}       % blackboard math symbols
\usepackage{nicefrac}       % compact symbols for 1/2, etc.
\usepackage{microtype}      % microtypography
\usepackage{xcolor}         % colors

\usepackage{graphicx}
\usepackage{amsmath}
\usepackage{amsthm}

\usepackage{booktabs}
\usepackage{algorithm}
\usepackage{algorithmic}

\title{You Only Need Half: Boosting Data Augmentation by Using Partial Content}

\author{%
  Juntao Hu \\
  School of Artificial Intelligence\\
  Jilin University\\
  Changchun, Jilin, China \\
  \texttt{jthu21@mails.jlu.edu.cn} \\
  \And Yuan Wu\thanks{Corresponding author} \\
  School of Artificial Intelligence\\
  Jilin University\\
  Changchun, Jilin, China \\
  \texttt{yuanwu@jlu.edu.cn} \\
}

\begin{document}

\maketitle

\begin{abstract}
  We propose a novel data augmentation method termed \textbf{Y}ou \textbf{O}nly \textbf{N}eed h\textbf{A}lf (YONA), which simplifies the augmentation process. YONA bisects an image, substitutes one half with noise, and applies data augmentation techniques to the remaining half. This method reduces the redundant information in the original image, encourages neural networks to recognize objects from incomplete views, and significantly enhances neural networks' robustness. YONA is distinguished by its properties of parameter-free, straightforward application, enhancing various existing data augmentation strategies, and thereby bolstering neural networks' robustness without additional computational cost. To demonstrate YONA’s efficacy, extensive experiments were carried out. These experiments confirm YONA's compatibility with diverse data augmentation methods and neural network architectures, yielding substantial improvements in CIFAR classification tasks, sometimes outperforming conventional image-level data augmentation methods. Furthermore, YONA markedly increases the resilience of neural networks to adversarial attacks. Additional experiments exploring YONA's variants conclusively show that masking half of an image optimizes performance. The code is available at \url{https://github.com/HansMoe/YONA}.
\end{abstract}

\section{Introduction}

Deep neural networks (DNNs) have achieved remarkable success across a variety of computer vision (CV) tasks, including image classification~\cite{paper01}, object detection~\cite{paper03}, and semantic segmentation~\cite{paper05}. Despite these advances, DNNs with intricate architectures tend to overfit extensive training datasets~\cite{russakovsky2015imagenet}, which compromises their generalization capabilities on unseen data. Additionally, DNNs are vulnerable to adversarial attacks~\cite{hendrycks2021natural,madry2018towards}, which further challenge their robustness. To mitigate these issues, numerous strategies have been developed, primarily focusing on enhancing DNNs' generalization and robustness. These strategies are broadly categorized into two groups: (1) regularization techniques~\cite{moosavi2019robustness,zhao2020domain} and (2) data augmentation techniques~\cite{rebuffi2021data,rebuffi2021fixing,zhou2024survey}. In this paper, we concentrate on exploring and expanding data augmentation methods to bolster the robustness of DNNs.

\begin{figure*}[t]
  \centering
  \includegraphics[width=0.8\textwidth]{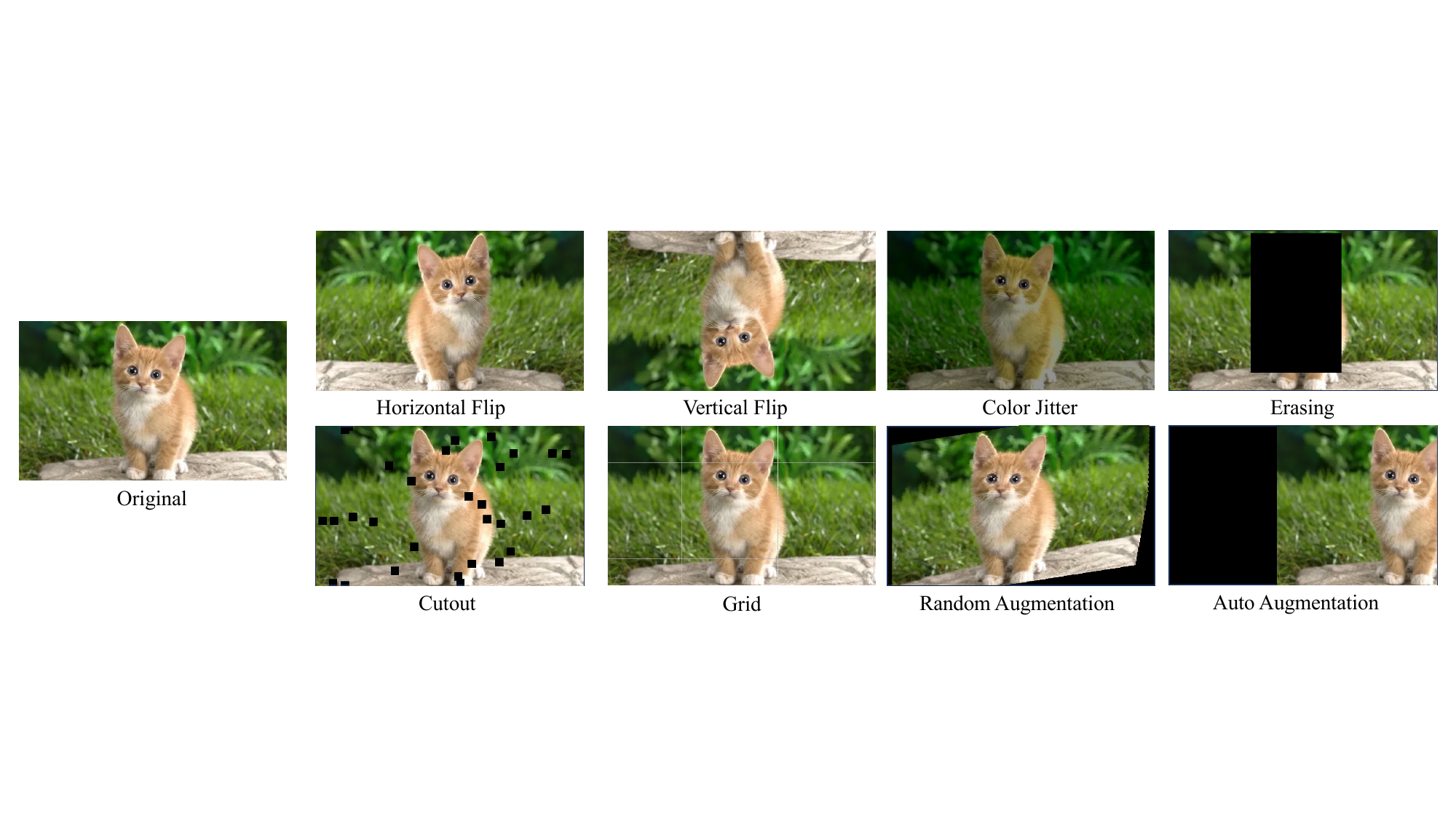}
  \caption{The illustrations of the studied data augmentation approaches.}
  \label{fig:aug}
\end{figure*}

Data augmentation techniques have become a focal point in enhancing the performance and robustness of DNNs in recent years~\cite{DASurvey01} (As shown in Fig.~\ref{fig:aug}). Drawing inspiration from human cognitive processes, these techniques predominantly operate at the image level, ensuring the preservation of global semantic integrity~\cite{DASurvey02}. Such methods aim to diversify the representation patterns of data, thereby narrowing the generalization gap observed between training and testing datasets. Notably, humans' ability to recognize objects from partial visuals suggests that image patches constitute potent natural signals. The Vision Transformer (ViT) has underscored this perspective by advocating the processing of patch sequences, positing this approach as the forthcoming paradigm within the CV community~\cite{dosovitskiy2020image}. ViT deconstructs an image into a series of non-overlapping patches, achieving results on par with traditional DNNs. Furthermore, the Fast Language-Image Pre-training (FLIP) technique introduces an innovative method by extensively masking and omitting large sections of image patches during training. This strategy demonstrates that masking enables DNNs to engage with an increased number of image-text pairs within the same temporal constraints, enhancing the ability to contrast more samples per iteration without expanding the memory requirements~\cite{li2023scaling}.

Inspired by ViT and FLIP, we hypothesize that executing data augmentation at the patch level is both feasible and beneficial. Conceptualizing an image as an assembly of multiple patches, or even as bifurcated sections, allows for the application of data augmentation techniques to these distinct segments independently. Furthermore, images often harbor considerable redundant information, some of which might adversely impact the training of DNNs~\cite{wang2020glance}. For a specific data augmentation, we could perform this method to transform certain pieces of an image, masking other pieces with noise, and subsequently reassembling the transformed and masked pieces into a unified image. Such a strategy aims to enrich patch-level diversity and diminish redundant information at the holistic image level, thereby fostering a DNN's ability to discern objects from incomplete visual information, akin to human perceptual capabilities.

Following the above hypothesis, we propose\textbf{Y}ou \textbf{O}nly \textbf{N}eed h\textbf{A}lf (YONA), a straightforward approach of performing data augmentations. YONA methodically bisects an image into two equal segments along either the vertical or horizontal axis. Subsequently, it applies data augmentation to one segment, masks the other with noise, and finally, reassembles the two altered segments into a cohesive image. This technique enhances DNN's ability to learn meaningful representations from fragmented visual inputs.

We executed comprehensive experiments encompassing a variety of data augmentation techniques, neural network architectures, and datasets. Specifically, for image classification tasks on CIFAR-10 and CIFAR-100, we assessed YONA with six neural network architectures and eight data augmentation methods. Out of 48 evaluations, YONA enhanced performance in 33 and 34 instances, respectively. Remarkably, YONA can improve the efficacy of all data augmentation methods across different neural architectures. In terms of robustness against adversarial attacks, we tested YONA using ResNet18 along with eight data augmentation strategies on CIFAR-10 and CIFAR-100 under the Projected Gradient Descent (PGD) adversarial attack, where it markedly amplified the effectiveness of all methods on both datasets. Additionally, an ablation study on CIFAR-10 was performed to empirically validate the effectiveness of masking half of an image's content within the YONA framework.

\section{Related Works}

\subsection{Data Augmentation}

Data augmentation is a pivotal technique for training DNNs and enhancing their generalization capabilities. Its primary goal is to augment the sufficiency and diversity of training data, thereby aiding DNNs in circumventing overfitting issues. Data augmentation strategies are broadly classified into two categories: basic and advanced methods. Basic data augmentation techniques include image manipulation, image erasing, and image mixing~\cite{zhou2024survey}. Image manipulations involve straightforward transformations such as rotation, flipping, and cropping. These methods are directly applied to images and are relatively simple to implement~\cite{kaur2021data}. Image erasing techniques, such as the cutout method~\cite{devries2017improved}, involve the deletion of image sub-regions, substituting the erased pixels with constant or random values. The random erasing method randomly selects a rectangular region within an image, replacing its pixels with random values~\cite{zhong2020random}. Image mixing techniques blend two or more images or their sub-regions into a single image. Techniques like Mixup create convex combinations of images and their labels, promoting simpler linear behavior during training~\cite{zhang2017mixup}, while Manifold Mixup enhances hidden representations and decision boundaries by mixing at multiple network layers~\cite{verma2019manifold}. All these basic data augmentation techniques are performed at the image level.

Advanced data augmentation techniques are categorized into auto augment and feature augmentation. Auto augment strategies automate the search for effective augmentation policies to maximize performance, involving a search algorithm and a defined search space. AutoAugment consists of two parts: search algorithm and search space. The search algorithm aims to find the best policy regarding the highest validation accuracy and the search space contains many policies that detail various augmentation operations and magnitudes with which the operations are applied~\cite{cubuk1805autoaugment}. Challenges with AutoAugment include the extensive computational cost due to its reliance on reinforcement learning~\cite{wang2020deep}. Alternatives like Fast AutoAugment~\cite{lim2019fast} and RandAugment~\cite{cubuk2020randaugment} streamline the process by reducing the search complexity and computational expense. Feature augmentation operates within a learned feature space, applying sophisticated transformations to enhance model training. FeatMatch learns feature-based refinement and augmentation methods to produce a varied set of complex transformations~\cite{kuo2020featmatch}. Moment Exchange encourages the DNNs to utilize the moment information of latent features to perform implicit data augmentation~\cite{li2021feature}.

\subsection{Patches in Vision}

Patches have been widely used as strong signals for various CV tasks, ranging from classification~\cite{vit} to image-to-image translation~\cite{park2020contrastive}. The ViT adopts a transformer architecture for CV tasks, eschewing traditional convolutional neural networks (CNNs). ViT processes input as sequences of patches, achieving state-of-the-art results~\cite{vit}. Patches also play a crucial role in data augmentation techniques. For instance, Patch Gaussian applies a Gaussian blur to specific image patches~\cite{lopes2019improving}, while CutMix replaces a patch of one image with a patch from another~\cite{yun2019cutmix}. The You Only Cut Once (YOCO) method divides an image into two halves, applies distinct augmentations to each, and recombines them into a single image~\cite{han2022you}. Our proposed YONA method innovatively manipulates image data by randomly removing 50\% of the information to mitigate the adverse effects of redundancy, differing from traditional data augmentation methods that process the entire image.

\section{Method}

\begin{figure*}[t]
  \centering
  \includegraphics[width=0.8\textwidth]{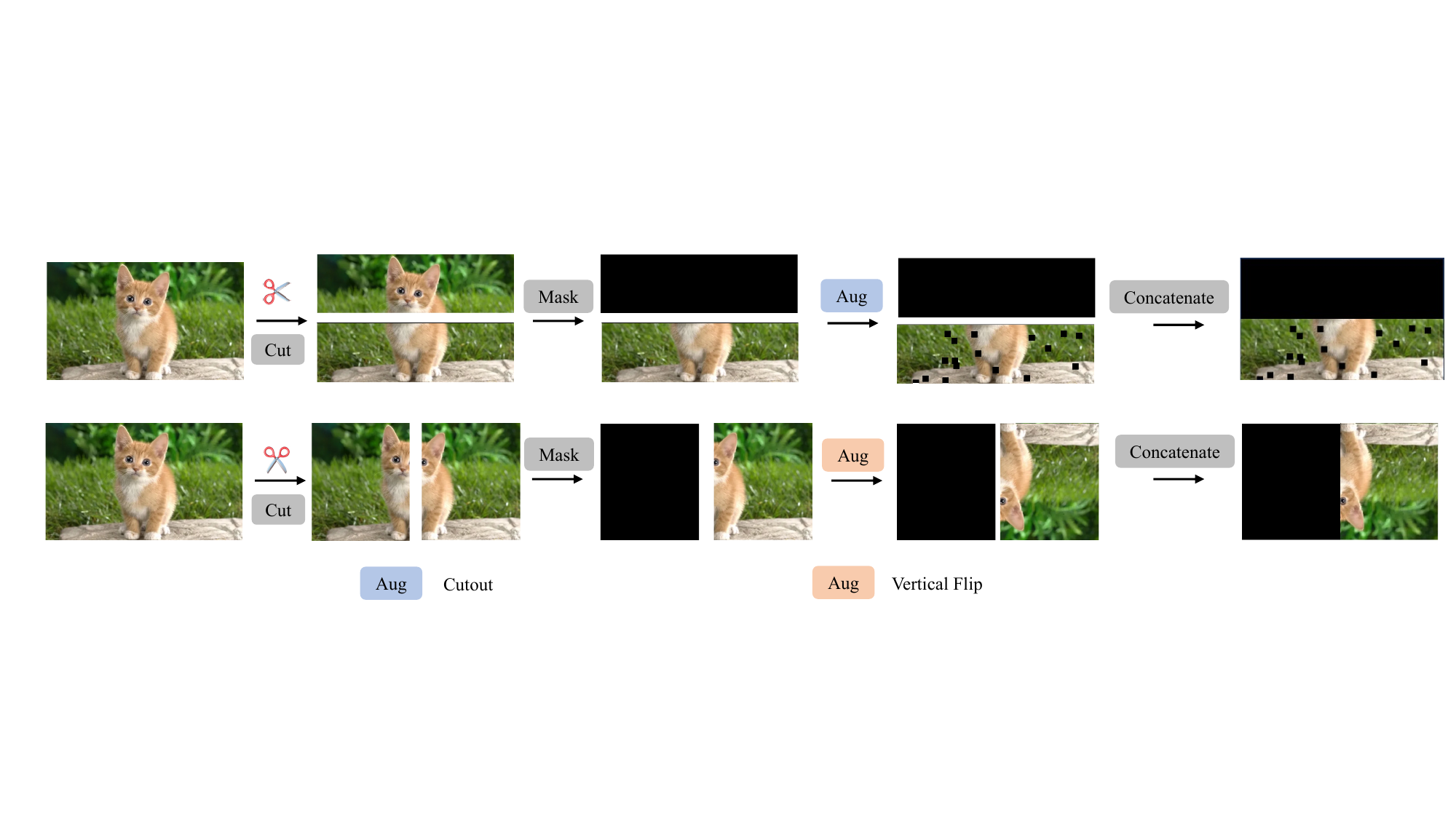}
  \caption{The illustrations of YONA operations. YONA cuts one image into two equal pieces, either in the height or the width dimension. A specific data augmentation method is performed on one piece, and the pixels within the other piece are replaced with noise. Transformed pieces are then concatenated together to form one single augmented image. The upper row shows the results of YONA applied with Cutout. The lower row presents the result of employing YONA to Vertical Flip.}
  \label{fig:short}
\end{figure*}

In this section, we delineate our proposed YONA method in detail. Consider an image $x \in \mathbb{R}^{C\times H\times W}$ and we let $a(\cdot)$ represent various data augmentation functions defined as $a(\cdot):\mathbb{R}^{C\times H\times W}\to\mathbb{R}^{C\times H\times W}$. Here, $x'=a(x)$ denotes the augmented image. Unlike conventional image-level data augmentation methods that directly apply augmentations to the entire image as $x'=a(x)$, YONA initially bifurcates the image into two equal segments along either the height or width dimension, each selection occurring with equal probability:

\begin{equation}
    \begin{split}
        \begin{aligned}
            \left[x_1,x_2\right]=cut_H(x),\quad if\quad0\leq p\leq0.5, \\ s.t.\quad x_1,x_2\in\mathbb{R}^{C\times\frac{H}{2}\times W}
        \end{aligned}
    \end{split}
  % \left[x_1,x_2\right]=cut_H(x),\quad if\quad0 \leq p \leq 0.5, \notag \\ s.t. \quad x_1,x_2 \in \mathbb{R}^{C\times H/2\times W}
\end{equation}
or
\begin{equation}
    \begin{split}
        \begin{aligned}
            \left[x_1,x_2\right]=cut_W(x),\quad if\quad 0.5\leq p\leq 1, \\ s.t. \quad x_1,x_2\in\mathbb{R}^{C\times H\times\frac{W}{2}}
        \end{aligned}
    \end{split}
  % \left[x_1,x_2\right]=cut_W(x),\quad if\quad0.5 < p \leq 1, \quad s.t. \quad x_1,x_2 \in \mathbb{R}^{C\times H\times W/2},
\end{equation}

\noindent where $p$ is sampled from $(0,1)$ uniformly (i.e., $p\in U(0,1)$), $x_1$ and $x_2$ are cut pieces, $cut_H(\cdot)$ and $cut_W(\cdot)$ represent the cut operation in the height dimension and the width dimension, respectively. Then $a(\cdot)$ is applied to one randomly selected piece, and the pixels within the other piece are replaced with noise as:

\begin{equation}
  x_1^M=mask(x_1) \quad and \quad x_2^A=a(x_2),if\quad0 \leq q \leq 0.5
\end{equation}
or
\begin{equation}
  x_1^A=a(x_1) \quad and \quad x_2^M=mask(x_2), if\quad 0.5 < q \leq 1,
\end{equation}

\noindent where $q$ is sampled from from $(0,1)$ uniformly (i.e., $q\in U(0,1)$), $x_1$ and $x_2$ are cut pieces, $x_1^M$ and $X_2^M$ are masked pieces, $x_1^A$ and $x_2^A$ are augmented pieces, $mask(\cdot)$ represents the masking with noise operation. Finally, we concatenate the transformed pieces back together as:

\begin{equation}
  x' = concat \left[x_1^M,x_2^A\right],
\end{equation}
or
\begin{equation}
  x' = concat \left[x_1^A,x_2^M\right],
\end{equation}

\noindent where $concat(\cdot,\cdot)$ represents the concatenation operation. Augmentation and masking operations are governed by randomness, encompassing random probabilities of being applied, random operations, and random magnitudes. These operations exhibit distinct behaviors, thereby instilling substantial diversity at both the patch and image levels. The overall flow of the YONA method is shown in Fig.~\ref{fig:short}.

\section{Experiments}

\subsection{Image Classification}

In this section, we assess the efficacy of YONA in classification tasks, focusing on its impact across the CIFAR-10 and CIFAR-100 datasets. We examine YONA's performance among six distinct neural network architectures: five CNNs and one ViT. Additionally, we explore eight different data augmentation techniques: Horizontal Flip (HFlip), Vertical Flip (VFlip), Color Jitter (Jitter), Random Erasing (Erasing), Cutout, Grid, Random Augmentation (RandAug), and Auto Augmentation (AutoAug). The illustrations of these data augmentation methods are present in Fig.~\ref{fig:aug}. \textbf{All details of different data augmentation methods are provided in the Appendix} \\

\noindent {\bf Experimental setup.} We evaluated 5 CNN architectures: ResNet-18~\cite{resnet}, Xception~\cite{chollet2017xception}, DenseNet-121~\cite{densenet121}, ResNeXt-50~\cite{resnet}, WRN-28-10~\cite{wideresnet} and ViT~\cite{vit} on CIFAR-10 and CIFAR-100 under the same training recipe, following the training setting provided in Co-Mixup~\cite{kim2021co}. In detail, CIFAR-10 and CIFAR-100 contain 50,000 training and 10,000 testing colored images drawn from 10 and 100 classes, respectively. Models were trained for 300 epochs, with an initial learning rate of 0.1 decayed by a factor of 0.1 at epochs 100 and 200. SGD optimizer with a momentum of 0.9 and a weight decay of 0.0001 was used in training. We used a batch size of 100, and models are trained on 2 GPUs. The training images were randomly cropped to $32\times32$ resolution with zero padding followed by random horizontal flip. Other data augmentation methods were added based on this setup. \\

\begin{table*}[t]
\centering
\caption{Test top-1 accuracy rate on CIFAR-10. We performed evaluations with 6 model architectures and 8 data augmentations. 32 results out of 48 were improved. $\triangle$ shows the improvement or decrease in data augmentation performance after combining with YONA.}
\resizebox{0.8\columnwidth}{!}{
\begin{tabular}{ l| c | c | c | c | c | c | c | c }
\toprule
Model & HFlip & VFlip & Jitter & Erasing & Cutout & Grid & RandAug & AutoAug \\
\midrule
ResNet-18 & 94.78 & 93.05 & 93.31 & 94.55 & 94.99 & 91.43 & 93.11 & 94.42 \\
+YONA & 95.40 & 92.82 & 94.12 & 93.85 & 94.64 & 93.71 & 92.82 & 94.65 \\
$\triangle$ & 0.62 & -0.23 & 0.81 & -0.7 & -0.35 & 2.28 & -0.29 & 0.23\\
\midrule
Xception & 93.33 & 92.96 & 94.13 & 93.58 & 92.91 & 92.15 & 92.47 & 94.75\\
+YONA & 95.06 & 92.58 & 93.79 & 94.12 & 94.23 & 93.07 & 92.53 & 94.13 \\
$\triangle$ & 1.73 & -0.38 & -0.34 & 0.54 & 1.32 & 0.92 & 0.06 & -0.62\\
% densenet121 to
\midrule
DenseNet-121 & 93.80 & 93.18 & 93.67 & 94.02 & 93.04 & 92.31 & 93.16 & 94.19 \\
+YONA & 94.98 & 92.43 & 94.62 & 93.45 & 94.28 & 93.11 & 92.62 & 94.51 \\
$\triangle$ & 1.18 & -0.75 & 0.95 & -0.57 & 1.24 & 0.8 & -0.54 & 0.32 \\
% resnet50 to
\midrule
ResNeXt-50 & 94.52 & 92.86 & 93.55 & 94.35 & 93.46 & 92.34 & 91.95 & 94.19 \\
+YONA & 95.45 & 92.2 & 93.89 & 94.23 & 94.16 & 93.31 & 92.54 & 94.44 \\
$\triangle$ & 0.93 & -0.66 & 0.34 & -0.12 & 0.7 & 0.97 & 0.59 & 0.25 \\
% wideresnet to
\midrule
WRN-28-10 & 95.24 & 92.71 & 93.4 & 94.08 & 93.51 & 91.55 & 92.53 & 94.57 \\
+YONA & 95.42 & 93.38 & 94.38 & 93.69 & 94.35 & 93.72 & 92.73 & 94.98 \\
$\triangle$ & 0.18 & 0.67 & 0.98 & -0.39 & 0.84 & 2.17 & 0.2 & 0.41 \\
% pretrained-Vit to
\midrule
ViT & 99.06 & 98.68 & 98.88 & 99.01 & 98.95 & 98.99 & 99.05 & 98.78 \\
+YONA & 99.00 & 98.67 & 99.06 & 98.98 & 98.97 & 99.05 & 99.13 & 99.20 \\
$\triangle$ & -0.06 & -0.01 & 0.18 & -0.03 & 0.02 & 0.06 & 0.08 & 0.42 \\
\bottomrule
\end{tabular} 
}
\label{tab:1_cifar10}
\end{table*}

\begin{table*}[t]
\centering
\caption{Test top-1 accuracy rate on CIFAR-100. We performed evaluations with 6 model architectures and 8 data augmentations. 34 results out of 48 were improved. $\triangle$ shows the improvement or decrease in data augmentation performance after combining with YONA.}
\resizebox{0.8\columnwidth}{!}{
\begin{tabular}{ l| c | c | c | c | c | c | c | c }
\toprule
Model & HFlip & VFlip & Jitter & Erasing & Cutout & Grid & RandAug & AutoAug \\
\midrule
ResNet-18 & 75.42 & 71.42 & 72.49 & 72.37 & 72.4 & 70.26 & 71.11 & 73.42 \\
    +YONA & 74.91 & 71.43 & 74.67 & 72.47 & 72.95 & 70.75 & 71.95 & 73.93 \\
    $\triangle$ & -0.51 & 0.01 & 2.18 & 0.1 & 0.55 & 0.49 & 0.84 & 0.51 \\
    % xception
    \midrule
    Xception & 75.96 & 70.75 & 72.4 & 73.04 & 71.38 & 70.31 & 70.6 & 73.15 \\
    +YONA & 75.32 & 70.36 & 72.48 & 73.11 & 71.92 & 70.1 & 70.72 & 73.67 \\
    $\triangle$ & -0.64 & -0.39 & 0.08 & 0.07 & 0.54 & -0.21 & 0.12 & 0.52 \\
    % densenet121 to
    \midrule
    DenseNet-121 & 71.46 & 72.54 & 72.54 & 72.53 & 72.53 & 70.19 & 71.18 & 73.57 \\
    +YONA & 74.54 & 72.38 & 72.74 & 72.71 & 73.14 & 70.6 & 71.95 & 72.95 \\
    $\triangle$ & 3.08 & -0.16 & 0.2 & 0.18 & 0.75 & 0.41 & 0.77 & -0.62 \\
    % resnet50 to
    \midrule
    ResNeXt-50 & 75.33 & 71.38 & 72.72 & 72.45 & 73.08 & 71.12 & 71.63 & 73.22 \\
    +YONA & 75.29 & 71.12 & 72.84 & 72.93 & 73.37 & 71.12 & 71.63 & 73.22 \\
    $\triangle$ & -0.04 & -0.26 & 0.12 & 0.48 & 0.29 & 0.46 & -0.75 & 0.39 \\
    % wideresnet to
    \midrule
    WRN-28-10 & 74.74 & 71.51 & 72.41 & 73.36 & 71.34 & 70.03 & 71.05 & 73.42 \\
    +YONA & 75.16 & 70.64 & 72.47 & 72.84 & 72.26 & 71.13 & 71.64 & 72.79 \\
    $\triangle$ & 0.42 & -0.87 & 0.06 & -0.52 & 0.92 & 1.1 & 0.59 & -0.63 \\
    % pretrained-Vit to
    \midrule
    ViT & 92.72 & 91.58 & 92.94 & 93.09 & 91.85 & 92.69 & 93.07 & 92.96 \\
    +YONA & 92.74 & 91.59 & 92.61 & 92.59 & 92.68 & 92.73 & 93.41 & 93.05 \\
    $\triangle$ & 0.02 & 0.01 & -0.33 & -0.5 & 0.83 & 0.04 & 0.34 & 0.09 \\
\bottomrule
\end{tabular} 
}
\label{tab:2_cifar10}
\end{table*}

\noindent{\bf Results.} We report the best results based on the mean of three independent trials. Table \ref{tab:1_cifar10} details the performance on the CIFAR-10 dataset. For the ResNet-18 architecture, YONA enhances performance in 4 out of 8 data augmentation scenarios, notably increasing the metric by 2.28\% for Grid. In the case of the Xception network, YONA boosts performance in 5 out of 8 scenarios, with significant improvements of 1.73\% for Horizontal Flip and 1.32\% for Cutout. With DenseNet-121, enhancements are noted in 5 augmentation methods. For ResNeXt-50, improvements are recorded in 6 out of 8 methods. The experimental results on WRN-28-10 show enhancements in 7 out of 8 methods, with a notable 2.17\% increase for Grid. For the ViT, improvements are observed in 5 out of 8 scenarios. Table \ref{tab:2_cifar10} displays results for CIFAR-100, where YONA enhances 34 out of 48 metrics. Notably, YONA consistently enhances performance for Grid on CIFAR-10 and for Cutout on CIFAR-100. Overall, YONA demonstrates superior generalization across diverse augmentation methods, with meaningful improvements (average improvement $\ge 0.2\%$) on both CIFAR-10 and CIFAR-100. Furthermore, the computational overhead and memory demands of YONA are negligible, matching the execution speed of standard image-level augmentations. This efficiency allows YONA to freely enhance multiple data augmentation methods across various neural network architectures.

\subsection{Object Detection}

In this section, we assess the efficacy of YONA in the object detection task, a complex scenario where the goal is to identify and pinpoint the locations and categories of objects within images. Unlike simple image classification, object detection necessitates not only recognizing objects but also precisely delineating their spatial presence. This task has inspired a variety of architectural and algorithmic innovations, among which certain models have proven particularly effective. Models such as the Faster Region-based Convolutional Neural Network (Faster R-CNN)~\cite{ren2015faster}, You Only Look Once (YOLO)~\cite{redmon2016you}, and Single Shot Multi-Box Detector (SSD)~\cite{liu2016ssd} have notably advanced the field, striking an optimal balance between accuracy and processing speed. In this investigation, we explore the impact of YONA on the YOLOv7 architecture~\cite{wang2023yolov7}, applying four different data augmentation methods across four evaluative metrics. \\

\noindent{\bf Experimental setup.} We trained the YOLOv7 architecture on the MS-COCO dataset~\cite{lin2014microsoft}, utilizing four evaluation metrics. The MS-COCO dataset, a well-established benchmark for object detection, comprises over 118,000 labeled images and 850,000 labeled object instances across 80 categories. Adhering to the standard object detection training protocol outlined in~\cite{wang2023yolov7}, we conducted training sessions for YOLOv7 on MS-COCO from scratch without leveraging pre-trained models. \textbf{Detailed training parameters are delineated in the Appendix}. We assessed the effectiveness of our YONA method alongside four data augmentations-Horizontal Flip (HFlip), Color Jitter (Jitter), Random Erasing (Erasing), and Random Augmentation (RandAug)—across four metrics: Precision (P), Recall (R), Mean Average Precision with Intersection over Union threshold above 0.5 (mAP@0.5), and Mean Average Precision with Intersection over Union thresholds ranging from 0.5 to 0.95 in increments of 0.05 (mAP@0.5:0.95). \\

\noindent\textbf{Results.} The experimental results in Table \ref{tab:3_yolo7} show that when applying YONA together with various data augmentations to tackle object detection, the results are limited. For three metrics: precision, mAP@0.5, and mAP@0.5:0.95, YONA brings negative effects. For recall, YONA improves the model's performance with three data augmentations: HFlip, Jitter, and RandAug. \\

\begin{table}[t]
\centering
\caption{Results of the object detection task. We evaluated four metrics with 4 data augmentations.}
\resizebox{0.5\columnwidth}{!}{
\begin{tabular}{ l| c | c | c | c }
\toprule
Method & P & R & mAP@0.5 & mAP@0.5:0.95 \\
    % HFlip
    \midrule
    HFlip & 0.604 & 0.533 & 0.558 & 0.391 \\
    +YONA & 0.585 & 0.547 &  0.539 & 0.335 \\
    $\triangle$ & -0.019 & \textbf{0.014} & -0.019 & -0.056 \\
    % Jitter
    \midrule
    Jitter & 0.721 & 0.635 & 0.685 & 0.493\\
    +YONA & 0.717 &  0.643 &  0.673 & 0.472 \\
    $\triangle$ & -0.004 & \textbf{0.008} & -0.012 &  -0.021\\
    % Erasing to
    \midrule
    Erasing & 0.724 & 0.624 & 0.68 & 0.49\\
    +YONA & 0.718 & 0.621 & 0.66 & 0.468\\
    $\triangle$ & -0.006 &  -0.003 & -0.02 & -0.022\\
    % RandomAug to
    \midrule
    RandAug & 0.731 &  0.603 & 0.679 & 0.51\\
    +YONA & 0.729 & 0.61 & 0.673 & 0.503 \\
    $\triangle$ & -0.002 & \textbf{0.007} & -0.006 & -0.007 \\
\bottomrule
\end{tabular} 
}
\label{tab:3_yolo7}
\end{table}

\noindent\textbf{Analyze.} Here we explore why YONA excels in classification tasks yet underperforms in object detection tasks. In classification, YONA's partial masking of an image allows the model to mitigate the detrimental effects of redundant information by focusing on the remaining content. This approach underscores that most information in one image is redundant or even harmful to downstream tasks, such as classification. Without reliance on self-supervised techniques to reconstruct the entire image from fragments, YONA effectively generalizes across various neural networks and data augmentations in classification. Conversely, object detection requires precise localization and identification of targets within an image. YONA's approach to masking half the image content risks removing significant portions of the target, potentially leading to incomplete or incorrect target recognition and localization. Furthermore, object detection algorithms often rely on contextual information surrounding targets to enhance accuracy; YONA's masking operation can strip away this context, impairing the model's ability to interpret semantic elements crucial for accurate detection. Additionally, self-supervised learning strategies are pivotal in object detection. The masked autoencoder (MAE)~\cite{he2022masked} has shown that masking up to 70\% of an image's patches can still yield effective results, highlighting the potential of self-supervised learning to compensate for extensive data masking. Lastly, masking may unevenly distribute categories within the dataset. If masking frequently occludes or erases objects from specific categories, the model may encounter fewer examples of these categories during training, thus degrading its ability to detect such categories effectively.

\subsection{Adversarial Robustness}

Adversarial robustness pertains to the resilience of neural networks against variations or deliberate perturbations in input data. In practical scenarios, several factors can compromise the integrity of input data, including changes in lighting, presence of noise, occlusion, and adversarial attacks~\cite{zhang2021survey}. Models trained exclusively on pristine, unaltered data are prone to vulnerabilities when exposed to noise and targeted attacks, often resulting in compromised performance. Selecting an effective data augmentation strategy is critical to developing a model with enhanced robustness. In this section, we assess the robustness of our newly developed YONA method against the PGD attack~\cite{gupta2018cnn}. \\

\noindent\textbf{PGD adversarial attack.} The PGD adversarial attack is a prevalent method within the CV community for assessing the robustness of CNN models. This iterative method crafts adversarial examples by leveraging gradient information of the input data, thereby misleading the model into incorrect classifications. PGD adversarial training, a robustness-enhancing strategy, fortifies neural networks against such attacks by integrating adversarial examples during training, thus enhancing the model's defensive capabilities. \\

\noindent{\bf Experimental setup.} Our experiments utilized the CIFAR-10 and CIFAR-100 datasets. We trained the ResNet-18 and employed PGD adversarial training strategy with the epsilon value of $8/255$ for $L_{\infty}$ bound, epsilon value of 0.25 (for attacking) or 0.5 (for training) for $L_2$ bound. For all attacks, PGD is with 4 steps of optimization. We compared these results against a baseline model trained without the YONA method. We borrow the implementation from Torchattacks~\cite{kim2020torchattacks}. The training configuration includes a batch size of 128, a weight decay of 0.0002, and a momentum of 0.9, spanning 200 epochs with a learning rate schedule of 0.1 for epochs 0 to 99, 0.01 for epochs 100 to 149, and 0.001 for epochs 150 to 200, following the protocols outlined in~\cite{madry2018towards}. \\

\noindent\textbf{Results.} Table \ref{tab:4_roubust_cifar10} presents the robust accuracy for the CIFAR-10 dataset, where YONA was tested with eight data augmentation techniques: Horizontal Flip (HFlip), Vertical Flip (VFlip), Color Jitter (Jitter), Random Erasing (Erasing), Cutout, Grid, Random Augmentation (RandAug), and Auto Augmentation (AutoAug). Notably, YONA significantly enhances robust classification accuracy under PGD attacks, showing an average improvement of 4.17\%. Specifically, enhancements with Jitter and Cutout are particularly substantial, each exceeding 5\%. Additionally, Table \ref{tab:5_roubust_cifar100} details the robust accuracy on the CIFAR-100 dataset, where YONA consistently boosts performance across all augmentation techniques, achieving an average accuracy increase of 4.21\%. Remarkably, the improvement attributed to the Cutout technique reaches 6.76\%. \textbf{We provide the experimental results under the Fast Gradient Sign Method (FGSM)~\cite{goodfellow2014explaining} in the Appendix.} \\

\begin{table*}[t]
\centering
\caption{Robust accuracy on CIFAR-10. We evaluated the robust accuracy of 8 data augmentation methods under the PGD adversarial attack with the ResNet-18 architecture.}
\resizebox{0.8\columnwidth}{!}{
\begin{tabular}{ l| c | c | c | c | c | c | c | c }
\toprule
Method & HFlip & VFlip & Jitter & Erasing & Cutout & Grid & RandAug & AutoAug \\
    % resnet18
    \midrule
    baseline & 47.16 & 42.94 & 47.25 & 50.36 & 46.65 & 47.22 & 50.37 & 51.39 \\
    +YONA & 51.06 & 49.24 & 52.38 & 51.88 & 52.92 & 51.42 & 53.62 & 54.17 \\
    $\triangle$ & 3.9 & 6.3 & 5.13 & 1.52 & 6.27 & 4.2 & 3.25 & 2.78\\
\bottomrule
\end{tabular} 
}
\label{tab:4_roubust_cifar10}
\end{table*}

\begin{table*}[t]
\centering
\caption{Robust accuracy on CIFAR-100. We evaluated the robust accuracy of 8 data augmentation methods under the PGD adversarial attack with the ResNet-18 architecture.}
\resizebox{0.8\columnwidth}{!}{
\begin{tabular}{ l| c | c | c | c | c | c | c | c }
\toprule
Method & HFlip & VFlip & Jitter & Erasing & Cutout & Grid & RandAug & AutoAug \\
    % resnet18
    \midrule
    baseline & 21.9 & 10.16 & 21.86 & 23.26 & 20.94 & 20.17 & 25.66 & 24.19 \\
    +YONA & 26.51 & 16.92 & 25.87 & 26.98 & 24.37 & 24.79 & 28.61 & 27.83 \\
    $\triangle$ & 4.61 & 6.76 & 4.01 & 3.72 & 3.43 & 4.62 & 2.95 & 3.64 \\
\bottomrule
\end{tabular} 
}
\label{tab:5_roubust_cifar100}
\end{table*}

\noindent\textbf{Analyze.} YONA achieves huge success in the robustness evaluation task against adversarial attacks, we summarize the key attributes as follows:

\begin{itemize}
    \item Random Perturbation: YONA introduces perturbations by randomly bisecting the image into two halves, masking one segment while applying data augmentations to the other. This procedure not only embeds noise into the image but also eliminates superfluous details, thereby enhancing model robustness. The inclusion of randomness enables the model to acquire features resilient to various perturbations, which consequently augments robust accuracy against adversarial attacks.
    \item Data Diversity: By randomly masking pixels across a sequence of image patches, YONA substantially enriches the diversity of the training data. This augmentation can bolster the model's resilience to adversarial threats and enhance its generalization capabilities. With YONA, models are exposed to a broader spectrum of feature representations, significantly reinforcing their robustness.
    \item Visual Invariance Learning: YONA compels models to discern and prioritize critical visual elements, mitigating the impact of extraneous information. The masking technique demands that the model concentrate on pivotal visual features, disregarding the irrelevant ones. This aspect of YONA equips the model with enhanced capabilities to withstand adversarial manipulations.
    \item Reduced Misleading Signals: By segmenting, transforming, and reassembling images, YONA ensures that adversarial samples generated from these modified instances are stripped of many misleading cues. This reduction in redundant information means that the adversarial examples are more instructive and less deceptive, thereby endowing the trained model with superior robustness.
\end{itemize}

\subsection{Ablation Study}

\begin{table}[t]
\centering
\caption{The results of the ablation study. We evaluated classification accuracy on CIFAR-10 of 4 data augmentations and two distinct masking rates with the ResNet-18 architecture.}
\resizebox{0.5\columnwidth}{!}{
\begin{tabular}{ l| c | c | c | c }
\toprule
 & HFlip & Jitter & Erasing & Grid\\
    % resnet18
    \midrule
    mask 1/4 & 94.82 & 92.53 & 93.25 & 92.86  \\
    mask 3/4 & 94.86 & 92.82 & 93.38 & 92.58 \\
    \midrule
    YONA & \textbf{95.4} & \textbf{94.12} & \textbf{93.45} & \textbf{93.71} \\
\bottomrule
\end{tabular} 
}
\label{tab:6_Ablation_cifar10}
\end{table}

YONA employs a simple yet effective approach to augment data augmentation techniques. It involves bisecting an image into two halves, applying a data augmentation to one half, masking the other with noise, and then merging the two pieces. It is critical that the masking rate is maintained at 50\% to ensure effectiveness. This study explores the impact of varying masking rates on the classification accuracy using the ResNet-18 architecture on the CIFAR-10 dataset. We assess four augmentation techniques: Horizontal Flip, Jitter, Erasing, and Grid, with two masking proportions: $1/4$ and $3/4$. \\

\noindent \textbf{Result.}  Table \ref{tab:6_Ablation_cifar10} displays the outcomes of the ablation study. The findings suggest that a 50\% masking rate consistently outperforms other rates across all data augmentation methods examined. Both alternative masking rates yielded suboptimal results, illustrating that increased masking can strip away crucial image information, while reduced masking may retain excessive redundancy. \textbf{More extensive experiments, including calibration, corruption robustness, and contrastive learning are provided in the Appendix to present more insights of YONA.}

\section{Conclusion}

In this paper, we introduce a simple, efficient, and straightforward method named \textbf{Y}ou \textbf{O}nly \textbf{N}eed h\textbf{A}lf (YONA) to enhance data augmentation techniques. YONA has demonstrated exceptional performance across various computer vision tasks, neural network architectures, and datasets. The efficacy of our YONA method underscores the critical role of data augmentations in training neural networks, particularly in developing robust models. Additionally, our findings highlight that a significant portion of the information in an image is superfluous; utilizing only partial information during training can lead to more efficient and robust models. We anticipate that our study will encourage the community to reevaluate and reconsider the application of data augmentation strategies.

\bibliographystyle{unsrt}  
\bibliography{ref} %++

\appendix

\section{Augmentations}

For all studied data augmentations, we list a brief description and implementation details here. For YONA, we do not modify the setting of augmentation operations (probability, magnitudes, etc.) unless specified. We employ the Pytorch/Torchvision implementation for most augmentations. \\

\noindent\textbf{Horizontal Flip} flips the image along the vertical central axis, that is, symmetrically transforms the image from left to right. We apply the Horizontal Flip with a probability of 0.5 as our default training setting. \\

\noindent \textbf{Vertical Flip} flips the image along the horizontal central axis, that is, symmetrically transforms the image from top to bottom. We apply the Vertical Flip with a probability of 0.5. \\

\noindent\textbf{Color Jitter} is an operation that performs color perturbation on an image, increasing the diversity of the data by randomly transforming the color channels of the image. These random transformations can include changing brightness, contrast, saturation, hue, etc. By applying Color Jitter, the model can be more robust to color changes. We set brightness = 0.4, contrast = 0.4, saturation = 0.4, and hue = 0.1. The probability of being applied is 0.5. \\

\noindent\textbf{Random Erasing} randomly selects a rectangle area in the image and replaces its pixels with random values. In this way, we can generate training images with various levels of occlusion, which reduces the risk of over-fitting and improves the robustness of models. Random Erasing is parameter learning free and easy to implement. We employ scale = (0.02, 0.4), ratio = (0.3, 3.3), and value = 0. Random Erasing is applied with a probability of 0.5. \\

\noindent\textbf{Cutout} randomly masking out square regions of input during training. It can be used to improve the robustness and overall performance of convolutional neural networks. The mask size for Cutout is set to 25\% of the image size and the location for dropping out is uniformly sampled. The dropping pixel is filled with 0 pixel. The probability of being applied is 0.5. \\

\noindent\textbf{Grid} aims to divide the input image into non-overlapping grids and apply one or more transformation operations to each grid. These transformations can include rotation, scaling, translation, shearing, etc. the probability of being applied is 0.5. \\

\noindent\textbf{Auto Augmentation} designed a search space in which a policy consists of a series of sub-policies, one of which is randomly chosen in each mini-batch. A sub-policy consists of two operations, each operation is an image processing method such as shearing, rotation, or translation, and the probabilities and magnitudes with which the functions are applied. \\

\noindent\textbf{Random Augmentation} consists of searched augmentations from a reduced search space, where the magnitude strength of augmentations is controllable. Random Augmentation can be applied uniformly across different tasks. In our experiments, we employ an identical setting: 2 augmentation operations and a magnitude of 9.

\section{Object Detection}

\subsection{Implementation Details}

We train the YOLOv7 architecture on the MS-COCO dataset. Models are trained for 300 epochs. The batch size is set to 16. The initial learning rate is 0.01. We use the Adam optimizer with a momentum of 0.937 and a weight decay of 0.0005. An additional training hyper-parameter is the top k of simOTA. To train $640\times640$ models, we follow YOLOX to use k = 10~\cite{wang2023yolov7} . All our experiments do not require pre-trained models, we train models from the scratch. 

\section{Adversarial Robustness}

We also employ the Fast Gradient Sign Method (FGSM)~\cite{goodfellow2014explaining} to verify the robustness of image classifiers. We borrow the implementation from Torchattacks~\cite{kim2020torchattacks}. We conduct experiments on both CIFAR-10 and CIFAR-100 and apply an $L_{\infty}$ budget of $8/255$. We follow the experimental protocol outlined by~\cite{han2022you}. To verify that our proposed YONA can be applicable to various architectures, here we examine VGG-19 ~\cite{simonyan2014very} with 8 data augmentations under the FGSM attack.

\begin{table*}[t]
\centering
\caption{Robust accuracy on CIFAR-10. We evaluated the robust accuracy of 8 data augmentation methods under the FGSM adversarial attack with the VGG-19 architecture.}
\resizebox{0.8\columnwidth}{!}{
\begin{tabular}{ l| c | c | c | c | c | c | c | c }
\toprule
Method & HFlip & VFlip & Jitter & Erasing & Cutout & Grid & RandAug & AutoAug \\
    % resnet18
    \midrule
    baseline & 9.11 & 8.90 & 9.14 & 8.52 & 8.29 & 9.63 & 9.03 & 9.08 \\
    +YONA & 9.70 & 9.17 & 9.31 & 8.92 & 8.67 & 11.53 & 9.16 & 9.19 \\
    $\triangle$ & 0.59 & 0.27 & 0.17 & 0.40 & 0.38 & 1.90 & 0.13 & 0.11 \\
\bottomrule
\end{tabular} 
}
\label{tab:1_roubust_cifar10}
\end{table*}

\begin{table*}[t]
\centering
\caption{Robust accuracy on CIFAR-100. We evaluated the robust accuracy of 8 data augmentation methods under the FGSM adversarial attacks with the VGG-19 architecture.}
\resizebox{0.8\columnwidth}{!}{
\begin{tabular}{ l| c | c | c | c | c | c | c | c }
\toprule
Method & HFlip & VFlip & Jitter & Erasing & Cutout & Grid & RandAug & AutoAug \\
    % resnet18
    \midrule
    baseline & 1.22 & 0.86 & 1.80 & 0.85 & 0.95 & 0.98 & 1.03 & 1.16 \\
    +YONA & 1.29 & 0.96 & 1.97 & 0.98 & 1.19 & 1.27 & 1.29 & 1.31 \\
    $\triangle$ & 0.07 & 0.10 & 0.17 & 0.13 & 0.24 & 0.29 & 0.26 & 0.15 \\
\bottomrule
\end{tabular} 
}
\label{tab:2_roubust_cifar100}
\end{table*}

\noindent\textbf{Results.} For robustness against the FGSM adversarial attack, we can observe that YONA can benefit all 8 data augmentations on both datasets from Table~\ref{tab:1_roubust_cifar10}\&\ref{tab:2_roubust_cifar100}. Therefore, we conclude that YONA can effectively enhance the robustness capability of neural networks.

\section{Calibration}

We follow the evaluation protocol used in PixMix~\cite{hendrycks2022pixmix}. We conduct the experiments on CIFAR-10 and CIFAR100 with a 40-4 Wide ResNet~\cite{zagoruyko2016wide} architecture under 8 data augmentations. The calibration task aims to match the empirical frequency of correctness. The posteriors from a model should satisfy

\begin{equation}
    \mathbb{P}(Y=arg\max_i f(X)_i|\max_i f(X)_i=C)=C
\end{equation}

\noindent where $f$ is an image classifier, $X, Y$ are random variables representing the data distribution. We use RMS calibration error~\cite{hendrycks2018deep} and adaptive binning~\cite{nguyen2015posterior}.

\begin{table*}[t]
\centering
\caption{Calibration results on CIFAR-10. We evaluated the RMS calibration error of 8 data augmentation methods with the 40-4 Wide ResNet architecture.}
\resizebox{0.8\columnwidth}{!}{
\begin{tabular}{ l| c | c | c | c | c | c | c | c }
\toprule
Method & HFlip & VFlip & Jitter & Erasing & Cutout & Grid & RandAug & AutoAug \\
    % resnet18
    \midrule
    baseline & 17.6 & 18.5 & 16.9 & 18.2 & 17.8 & 17.1 & 15.2 & 14.8 \\
    +YONA & 17.8 & 18.2 & 16.8 & 18.0 & 17.7 & 16.8 & 14.8 & 14.6 \\
    $\triangle$ & 0.2 & -0.3 & -0.1 & -0.2 & -0.1 & -0.3 & -0.4 & -0.2 \\
\bottomrule
\end{tabular} 
}
\label{tab:3_roubust_cifar10}
\end{table*}

\begin{table*}[t]
\centering
\caption{Calibration results on CIFAR-100. We evaluated the RMS calibration error of 8 data augmentation methods with the 40-4 Wide ResNet architecture.}
\resizebox{0.8\columnwidth}{!}{
\begin{tabular}{ l| c | c | c | c | c | c | c | c }
\toprule
Method & HFlip & VFlip & Jitter & Erasing & Cutout & Grid & RandAug & AutoAug \\
    % resnet18
    \midrule
    baseline & 31.2 & 31.0 & 29.9 & 31.0 & 31.1 & 29.8 & 25.4 & 24.9 \\
    +YONA & 31.0 & 30.9 & 29.5 & 31.3 & 30.9 & 29.7 & 25.6 & 24.6 \\
    $\triangle$ & -0.2 & -0.1 & -0.4 & 0.3 & -0.2 & -0.1 & 0.2 & -0.3 \\
\bottomrule
\end{tabular} 
}
\label{tab:4_roubust_cifar100}
\end{table*}

\noindent\textbf{Results.} In terms of calibration, we could observe that YONA can reduce the RMS calibration error for all data augmentations except for the Horizontal Flip on the CIFAR-10 dataset, and for all data augmentations except for the Random Erasing and Random Augmentation on the CIFAR-100 dataset from Table~\ref{tab:3_roubust_cifar10}\&\ref{tab:4_roubust_cifar100}, respectively. To conclude, YONA helps image classifiers to have better-calibrated predictions most of the time.

\section{Corrupt Robustness.} Following Co-Mixup~\cite{kim2021co}, we evaluate image classifiers on two challenging corrupted test sets, to verify the generalization ability and robustness on unseen environments. The corrupted test sets are created with the following operations: (1) replace the background with a random image and (2) add Gaussian noise to the background. We conduct experiments on CIFAR-10 and CIFAR-100 with ResNet-18 under 8 data augmentations.

\begin{table*}[t]
\centering
\caption{Corruption robustness results on CIFAR-10. We evaluated the accuracy of 8 data augmentation methods with the ResNet-18 architecture under the random replace strategy.}
\resizebox{0.8\columnwidth}{!}{
\begin{tabular}{ l| c | c | c | c | c | c | c | c }
\toprule
Method & HFlip & VFlip & Jitter & Erasing & Cutout & Grid & RandAug & AutoAug \\
    % resnet18
    \midrule
    baseline & 46.41 & 42.74 & 47.18 & 46.92 & 47.03 & 46.28 & 46.99 & 47.21 \\
    +YONA & 48.92 & 44.31 & 48.14 & 47.88 & 48.91 & 47.57 & 48.13 & 48.71 \\
    $\triangle$ & 2.51 & 1.57 & 0.96 & 0.96 & 1.88 & 1.29 & 1.14 & 1.50 \\
\bottomrule
\end{tabular} 
}
\label{tab:5_roubust_cifar10}
\end{table*}

\begin{table*}[t]
\centering
\caption{Corruption robustness results on CIFAR-100. We evaluated the accuracy of 8 data augmentation methods with the ResNet-18 architecture under the random replace strategy.}
\resizebox{0.8\columnwidth}{!}{
\begin{tabular}{ l| c | c | c | c | c | c | c | c }
\toprule
Method & HFlip & VFlip & Jitter & Erasing & Cutout & Grid & RandAug & AutoAug \\
    % resnet18
    \midrule
    baseline & 19.41 & 15.57 & 18.95 & 19.17 & 19.64 & 18.79 & 19.29 & 20.11 \\
    +YONA & 21.12 & 16.68 & 21.07 & 20.53 & 20.18 & 19.84 & 20.41 & 22.46 \\
    $\triangle$ & 1.71 & 1.11 & 2.12 & 1.36 & 0.54 & 1.05 & 1.12 & 2.35 \\
\bottomrule
\end{tabular} 
}
\label{tab:6_roubust_cifar100}
\end{table*}

\begin{table*}[t]
\centering
\caption{Corruption robustness results on CIFAR-10. We evaluated the accuracy of 8 data augmentation methods with the ResNet-18 architecture under the Gaussian noise strategy.}
\resizebox{0.8\columnwidth}{!}{
\begin{tabular}{ l| c | c | c | c | c | c | c | c }
\toprule
Method & HFlip & VFlip & Jitter & Erasing & Cutout & Grid & RandAug & AutoAug \\
    % resnet18
    \midrule
    baseline & 48.93 & 40.08 & 46.13 & 45.93 & 47.32 & 48.21 & 47.51 & 47.35 \\
    +YONA & 49.17 & 42.95 & 49.74 & 47.12 & 47.96 & 49.34 & 48.39 & 48.61 \\
    $\triangle$ & 0.24 & 2.87 & 3.61 & 1.19 & 0.64 & 1.13 & 0.88 & 1.26 \\
\bottomrule
\end{tabular} 
}
\label{tab:7_roubust_cifar10}
\end{table*}

\begin{table*}[t]
\centering
\caption{Corruption robustness results on CIFAR-100. We evaluated the accuracy of 8 data augmentation methods with the ResNet-18 architecture under the Gaussian noise strategy.}
\resizebox{0.8\columnwidth}{!}{
\begin{tabular}{ l| c | c | c | c | c | c | c | c }
\toprule
Method & HFlip & VFlip & Jitter & Erasing & Cutout & Grid & RandAug & AutoAug \\
    % resnet18
    \midrule
    baseline & 20.15 & 17.78 & 21.71 & 20.57 & 21.46 & 21.07 & 21.83 & 21.96 \\
    +YONA & 23.36 & 18.65 & 22.93 & 21.96 & 22.19 & 21.60 & 22.86 & 22.63 \\
    $\triangle$ & 3.21 & 0.87 & 1.22 & 1.39 & 0.73 & 0.53 & 1.03 & 0.67 \\
\bottomrule
\end{tabular} 
}
\label{tab:8_roubust_cifar100}
\end{table*}

\noindent\textbf{Results.} From Table~\ref{tab:5_roubust_cifar10}\&\ref{tab:6_roubust_cifar100}\&\ref{tab:7_roubust_cifar10}\&\ref{tab:8_roubust_cifar100}, it is noteworthy that YONA can consistently show better performance when evaluated with challenging corrupted images. 8 data augmentations are all boosted by YONA on both datasets. This suggests that when only partial information is available, YONA with a better ability to recognize partial objects behaves more robustly.

\section{Contrastive Learning}

One core step of contrastive learning is to create two views from one image by data augmentations. YONA increases the diversity of data, applying YONA to contrastive learning yields more challenging views. In this section, we verify whether YONCA benefits MoCo v2~\cite{chen2020improved} and SimSiam~\cite{chen2021exploring} in the pre-training stage. The learned embedding is evaluated via multiple downstream tasks, including VOC detection~\cite{everingham2010pascal}, COCO detection and COCO instance segmentation~\cite{lin2014microsoft}.

\noindent\textbf{Implementation Details.} We strictly follow the training and evaluation protocols shown in ~\cite{han2022you}. In this study, we apply YONA to Horizontal Flip only, pre-train MoCo v2 and SimSiam with YONA for 200 and 100 epochs respectively. We follow the augmentations used in MoCo v2 and SimSiam. The augmentation procedure includes: Random resized crop with scale $[0.2, 0.1]$, Color jitter with brightness = 0.4, contrast = 0.4, saturation = 0.4, hue = 0.1, and being applied with a probability of 0.5. Grayscale with a probability of 0.8, Gaussian blur with sigma = (0.1, 2.0) and is applied with a probability of 0.5. The backbone is ResNet-50. MoCo v2 is trained with SGD as the optimizer, using a weight decay of 0.0001 and a momentum of 0.9. The initial learning rate is 0.03, and the learning rate is multiplied by 0.1 at 120 and 160 epochs. The temperature is set to 0.2. SimSiam is trained with the SGD optimizer with a momentum of 0.9, the weight decay is 0.0001, and the learning rate is scaled to 0.05 with a cosine decay schedule. For the VOC detection, Faster R-CNN with the C4-backbone is fine-tuned on VOC 2007 trainval+2012 train and evaluated on the VOC 2007 test. We halve the batch size from 16 to 8, double the max iterations from 24K to 48K, and scale the base learning rate from 0.02 to 0.01. All reported results are the average over 3 trials. For the COCO detection and instance segmentation, the model is Mask R-CNN with a C4 backbone. Similar to VOC, batch size and base learning rate are halved while the maximum iteration is doubled, from 90K to 180K.

\noindent\textbf{Results.} Table~\ref{tab:9_roubust_cifar100}\&\ref{tab:10_roubust_cifar100}\&\ref{tab:11_roubust_cifar100} depict the results. Among multiple tasks and metrics, YONA can enhance most of them. In particular, YONA could boost the performance of COCO detection with both MoCo v2 and SimSiam networks.

\begin{table*}[t]
\centering
\caption{Results of VOC detection.}
\resizebox{0.3\columnwidth}{!}{
\begin{tabular}{ l| c | c | c }
\toprule
Method & $AP_{50}$ & $AP$ & $AP_{75}$ \\
    % resnet18
    \midrule
    MoCo v2 & 82.4 & 57.0 & 63.6 \\
    +YONA & 82.6 & 56.9 & 64.1 \\
    $\triangle$ & 0.2 & -0.1 & 0.5 \\
    \midrule
     SimSiam & 80.1 & 54.3 & 60.2 \\
    +YONA & 80.0 & 54.6 & 60.1 \\
    $\triangle$ & -0.1 & 0.3 & -0.1 \\
\bottomrule
\end{tabular} 
}
\label{tab:9_roubust_cifar100}
\end{table*}

\begin{table*}[t]
\centering
\caption{Results of COCO detection.}
\resizebox{0.3\columnwidth}{!}{
\begin{tabular}{ l| c | c | c }
\toprule
Method & $AP_{50}$ & $AP$ & $AP_{75}$  \\
    % resnet18
    \midrule
    MoCo v2 & 57.2 & 37.9 & 40.8 \\
    +YONA & 57.3 & 38.2 & 41.1 \\
    $\triangle$ & 0.1 & 0.3 & 0.3 \\
    \midrule
     SimSiam & 51.6 & 33.9 & 36.1 \\
    +YONA & 51.7 & 34.3 & 36.8 \\
    $\triangle$ & 0.1 & 0.4 & 0.7 \\
\bottomrule
\end{tabular} 
}
\label{tab:10_roubust_cifar100}
\end{table*}

\begin{table*}[t]
\centering
\caption{Results of COCO instance segmentation.}
\resizebox{0.3\columnwidth}{!}{
\begin{tabular}{ l| c | c | c }
\toprule
Method & $AP^{mask}_{50}$ & $AP$ & $AP^{mask}_{75}$  \\
    % resnet18
    \midrule
    MoCo v2 & 54.2 & 33.4 & 35.6 \\
    +YONA & 54.0 & 33.5 & 35.5 \\
    $\triangle$ & -0.2 & 0.1 & -0.1 \\
    \midrule
     SimSiam & 48.5 & 29.9 & 32.1 \\
    +YONA & 49.2 & 30.5 & 32.6 \\
    $\triangle$ & 0.7 & 0.6 & 0.5 \\
\bottomrule
\end{tabular} 
}
\label{tab:11_roubust_cifar100}
\end{table*}

%%%%%%%%%%%%%%%%%%%%%%%%%%%%%%%%%%%%%%%%%%%%%%%%%%%%%%%%%%%%

\end{document}